\documentclass[a4paper, 10pt, conference]{ieeeconf}      
\usepackage{FG2024}

\FGfinalcopy 

\IEEEoverridecommandlockouts                              
\usepackage{graphics} 
\usepackage{epsfig} 
\usepackage{mathptmx} 
\usepackage{times} 
\usepackage{amsmath} 
\usepackage{amssymb}  
\usepackage{hyperref}
\usepackage{subcaption}

\let\OLDthebibliography\thebibliography
\renewcommand\thebibliography[1]{
  \OLDthebibliography{#1}
  \setlength{\parskip}{0pt}
  \setlength{\itemsep}{0pt plus 0.3ex}
}

\def\FGPaperID{203} 

\title{\LARGE \bf
A Gloss-free Sign Language Production with Discrete Representation
}

\author{\parbox{16cm}{\centering
    {\large Eui Jun Hwang, Huije Lee, and Jong C. Park\(^\dagger\)}\\
    {\normalsize
    Korea Advanced Institute of Science and Technology (KAIST), Daejeon, Korea}}
    \thanks{\(^\dagger\)Corresponding author.}
}

\usepackage{fancyhdr}
\begin{document}

\ifFGfinal
\thispagestyle{empty}
\pagestyle{empty}
\else
\author{Anonymous FG2024 submission\\ Paper ID \FGPaperID \\}
\pagestyle{plain}
\fi
\maketitle
\thispagestyle{fancy}
\begin{abstract}

Gloss-free Sign Language Production (SLP) offers a direct translation of spoken language sentences into sign language, bypassing the need for gloss intermediaries. Previous autoregressive SLP methods have not fully achieved true autoregression, as they often depend on ground-truth data during inference. To fill this gap, we introduce Sign language Vector Quantization Network (SignVQNet), leveraging discrete spatio-temporal representations of sign poses. With such a discrete representation, our method incorporates beam search, a decoding strategy widely used in Natural Language Processing. Furthermore, we align the discrete representation with linguistic features from pre-trained language models such as BERT. Our results show the superior performance of our method over prior SLP methods in generating accurate and realistic sign pose sequences. Additionally, our analysis shows that the reliability of Back-Translation and Fr\'{e}chet Gesture Distance as evaluation metrics, in contrast to DTW-MJE. The code and models are available at \href{https://github.com/eddie-euijun-hwang/SignVQNet}{\textcolor{magenta}{https://github.com/eddie-euijun-hwang/SignVQNet}}.

\end{abstract}

\section{INTRODUCTION}

Gloss-free Sign Language Production (SLP) directly translates spoken language into sign poses, eliminating the need for gloss annotation. Glosses, while providing a direct mapping between spoken language and sign poses, require significant labor, time, and specialized knowledge of sign language. This high demand for the resources has been a driving factor in the growing interest and transition towards gloss-free methods \cite{li2020tspnet,lin-etal-2023-gloss,shi-etal-2022-open,Yin_2023_CVPR}. While the gloss-free methods typically exhibit lower performance than the gloss-based ones, they offer greater accessibility and efficiency. 

In the domain of gloss-free SLP, two approaches prevail: retrieval and generative models. Retrieval models \cite{Cheng_2023_CVPR,duarte2022sign,saunders2022signing} fetch relevant samples from datasets based on textual prompts. Generative models \cite{hwang2021non,saunders2020progressive,saunders2021mixed}, on the other hand, can generate entirely new signing sequences by leveraging patterns learned during training. This capability to produce diverse outputs makes generative models a compelling choice for SLP, which is the focus of our research. However, the generative models face a few challenges. The length disparity between sign pose sequences and their spoken equivalents often necessitates clustering sequences into gloss-level representations \cite{lin-etal-2023-gloss}. Moreover, the non-linear nature of sign language compared to the linear structure of spoken language adds complexity to this task.

Recent studies \cite{huang2021towards,hwang2021non} have pointed out the constraints of the model introduced earlier \cite{saunders2020progressive}. A significant concern is about its dependence on the initial ground-truth pose and timing for inference, pivotal for the model's autoregression. The continuous nature of the sign pose sequences, represented as keypoint data, complicates achieving true autoregression without auxiliary information during inference. 

\begin{figure}[t]
\centering
\includegraphics[width=0.9\columnwidth]{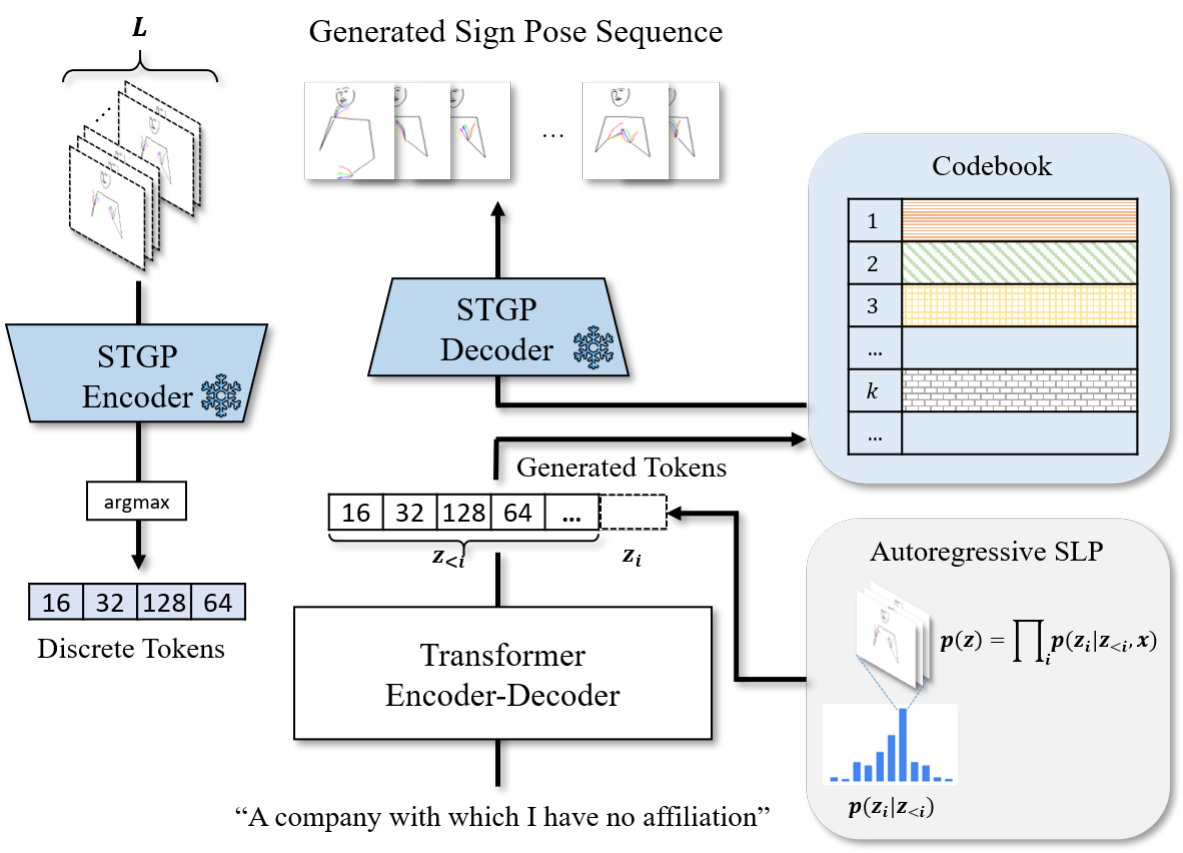} 
\caption{An overview of SignVQNet, where both the pre-trained STGP encdoer and decoder remain frozen. The encoder converts the sign pose sequence into discrete tokens. These tokens are then generated from textual inputs by Transformer. The decoder transforms the generated tokens back into an actual sign pose sequence.}
\vspace{-1.5em}
\label{model-overview}
\end{figure}

To address this, we introduce \textbf{Sign} language \textbf{V}ector \textbf{Q}uantization \textbf{Net}work (\textbf{SignVQNet}), as shown in Fig. \ref{model-overview}. Leveraging vector quantization, our method converts the sign pose sequences into discrete tokens, enabling genuine autoregressive generation. This approach also supports beam search, commonly used for Natural Language Processing (NLP) tasks \cite{cho-etal-2023-discrete,jeong-etal-2021-unsupervised,10193533}. Additionally, we introduce latent-level alignment to directly associate linguistic features with sign pose features. 

We compared the performance of SignVQNet against those of other existing SLP models using Back-Translation (BT) \cite{saunders2020progressive}, DTW-MJE \cite{huang2021towards}, and Fr\'{e}chet Gesture Distance (FGD) \cite{yoon2020speech}. Our experimental results showed that SignVQNet consistently outperformed the previous methods on two sign language datasets: RWTH-PHOENIX-WEATHER-2014T \cite{camgoz2018neural} and How2Sign \cite{duarte2021how2sign}. In our additional experiments focusing on beam size adjustment of our method, we found that DTW-MJE suffers from inconsistencies, raising questions about its reliability as a suitable metric for SLP. By contrast, both BT and FGD have demonstrated better consistency and reliability as more effective metrics for assessing SLP.

\section{RELATED WORKS}\label{sec:related}

\subsection{Generative Gloss-free Sign Language Production} 

Saunders et al. \cite{saunders2020progressive} pioneered the application of Progressive Transformers (PT) to gloss-free SLP. Their approach combined a counter decoding method and augmentation strategies such as Gaussian Noise and Future Prediction. Additionally, they introduced BT to assess the model performance. Building on this, Saunders et al. \cite{saunders2020adversarial} addressed the regression-to-the-mean issue by adopting an adversarial training framework. In their subsequent work, they optimized PT by employing a mixture of motion primitives \cite{saunders2021mixed}. Meanwhile, Hwang et al. \cite{hwang2021non} introduced a paradigm shift with their Non-Autoregressive Sign Language Production with a Gaussian space (NSLP-G) model. Designed to convert spoken language sentences into corresponding sign pose sequences, NSLP-G diverged from conventional methods by adopting non-autoregressive decoding with the pre-trained VAE on the spatial aspect of the sign pose sequences. In our work, we extend this exploration into the spatial-temporal aspect of sign pose sequences, aiming to achieve a gloss-level representation.

\subsection{Discrete Representation} 

There are several studies that convert continuous data into discrete data. Maddison et al. \cite{maddison2016concrete} and Jang et al. \cite{jang2016categorical} propose Concrete Distribution and Gumbel Softmax Relaxation, respectively, which are techniques for approximating the sampling process of discrete data from a continuous distribution using annealing during training. Van et al. \cite{van2017neural} propose VQ-VAE, which extends the standard autoencoder by adding a discrete codebook component to the network. VQ-VAE compares the vector in the codebook with the output of the encoder, where the closest vector is fed to the decoder. The model is trained using an online cluster assignment procedure coupled with a straight-through estimator. Gumbel Softmax Relaxation allows the model to effectively learn a discrete latent distribution \cite{ramesh2021zero}. In our work, we utilize this method to discretize the sign pose sequences.

\section{METHOD}

\subsection{Problem Formulation}

Consider a spoken language sentence $x=\{x_u\}^U_{u=1}$, which consists of $U$ words. The objective of SLP is to produce a sign pose sequence $y=\{y_t\}^T_{t=1}\in \mathbb{R}^{V\times C}$, where $V$ denotes the number of vertices, and $C$ the feature dimension of the skeletal pose data. Instead of directly modeling $p(y|x)$, we employ an intermediary representation $z$, consisting of discrete tokens. These tokens encapsulate both spatial and temporal attributes of sign language. The generation process is then defined by the joint distribution $p(y,z|x)=q(y|z,x)p(z|x)$. Here, $p(z|x)$ denotes the probability of generating the discrete representation $z$ from the input $x$, while $q(y|z,x)$ represents the probability of generating the continuous sign pose sequence $y$ based on $z$ and $x$. The first term is handled by a vector quantization model, and the second is modeled in an autoregressive manner. 

\begin{figure}[t!]
\centering
\includegraphics[width=1.0\columnwidth]{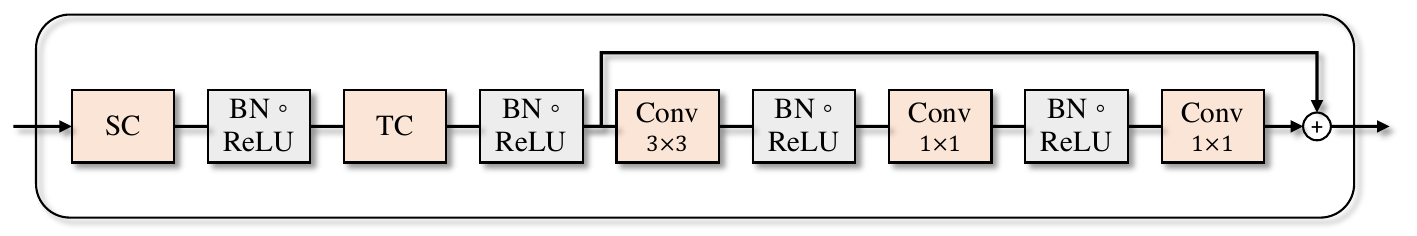}
\vspace{-1.5em}
\caption{An overview of the STGP block. Spatial Convolution (SC) and Temporal Convolution (TC) process the input spatially and temporally, respectively, and the subsampled output preserves spatio-temporal features through a residual connection. BN refers to Batch Normalization.}
\vspace{-1.5em}
\label{fig:stgp}
\end{figure}

\subsection{Learning Discrete Representation of Sign Poses}

To convert a sign pose sequence into discrete tokens, we introduce Spatio-Temporal Graph Pyramid (STGP)-based dVAE \cite{ramesh2021zero}, as shown in Fig. \ref{model-overview}. Inspired by the Graph Pyramid \cite{yan2019convolutional}, we propose an STGP block--a fundamental building block for the encoder and decoder--to address the intricacies of down-sampling and up-sampling within a skeleton graph, which features non-uniform grids. A detailed overview of the STGP block is provided in Fig. \ref{fig:stgp}. The STGP block sequentially processes input sequences, handling both their spatial and temporal aspects. Each processing ends with a residual connection to preserve down-sampled spatio-temporal information.

Our model is designed to handle fixed-length sign segments represented as $y^{(\ell)}\in \mathbb{R}^{L\times V\times C}$, where $L$ represents the window size. A sign segment refers to a small fraction of the full sign pose sequence. Central to the STGP-dVAE is a codebook comprising latent variable categories, represented as $e_i\in \mathbb{R}^{K\times D_c}$. Here, $K$ denotes the number of latent variable categories, while $D_c$ refers to the embedding size. Then we discretize the output of the STGP encoder using Gumbel-Softmax relaxation \cite{jang2016categorical}. This process samples a latent from the output of the encoder $e_i$ as:
\begin{equation}
    w_i=\frac{\exp(\log(e_i)+g_i)/\tau}{\sum_{k=1}^{K}\exp(\log(e_k)+g_k)/\tau},
\end{equation}
where $g_i$ represents the independent $i^{th}$ sample from the Gumbel distribution. The parameter $\tau$ adjusts the approximation to the categorical distribution, and $w_i$ denotes the weights over the codebook vectors. These discretized representations are used for ``sign tokens'' in subsequent training shown in Sec. \ref{sec:arslp}. The resulting sampled latent vector is then given by $z^{(\ell)}=\sum_{k=1}^{K}w_ke_k$.

The model is optimized by minimizing the combined loss function that consists of reconstruction and diversity losses \cite{baevski2020wav2vec}. We use L2 loss as the reconstruction loss, defined as:
\begin{equation}
    \mathcal{L}_{pose} = \frac{1}{L} \sum_{i=1}^{L} \left \| y^{(\ell)}_i - \hat{y}^{(\ell)}_i \right \|^2_2,
\end{equation}
where $y^{(\ell)}$ and $\hat{y}^{(\ell)}$ represent the ground-truth and the reconstructed sign segment, respectively. The diversity loss, which enables the model to use the codebook effectively \cite{baevski2020wav2vec}, is represented as:
\begin{equation}
    \mathcal{L}_{div} = -\sum_{i=1}^K p(e_i)\log{(p(e_i))}.
\end{equation}
The final loss can be defined as:
\begin{equation}
    \mathcal{L} = \mathcal{L}_{pose} + \alpha \mathcal{L}_{div},
\end{equation}
where $\alpha$ is the hyperparameter that determines the scale of diversity loss.

\subsection{Autoregressive Sign Language Production} \label{sec:arslp}

To generate the sign tokens from the given spoken language sentence, we use the Transformer encoder-decoder architecture, as depicted in Fig. \ref{model-overview}. The first step involves converting the sign pose sequence into the sign tokens. This process entails dividing the input sign pose sequence, $y$, into multiple sign segments, each with a  length $L$. Consequently, the number of sign tokens can be $M=\left \lfloor\frac{T}{L} \right \rfloor$, resulting in $y \approx \{y^{(\ell)}_i\}_{i=1}^{M}\in \mathbb{R}^{M\times V\times C}$. For computational efficiency, any remaining sign poses are simply removed. Each segment is subsequently encoded by the pre-trained STGP encoder through the argmax operation, yielding a sign token denoted as $z_i=\mathrm{argmax}(h_i)$, where $h_i$ is the $i^{th}$ hidden representation from the STGP encoder. To mark the start and end of signs, $z$ is padded with $\left \langle \mathrm{bos} \right \rangle$ and $\left \langle \mathrm{eos} \right \rangle$.

The model is optimized by minimizing the combined loss function, which consists of Cross-Entropy (CE) and latent alignment losses. The CE loss is represented as:
\begin{equation}
\mathcal{L}_{ce}=-\sum_{i=1}^{M}\log (p(z_i|z_{<i}, x)),
\end{equation}
where $p(z_i|z_{<i}, x)$ represents the probability of generating the $i^{th}$ token $z_i$ given the previous tokens $z_{<i}$ and the input $x$.

The latent loss, employing L2 loss, offers supplementary latent-level signals by aligning the output of the Transformer decoder with that of the pre-trained STGP encoder. It can be defined as:
\begin{equation}
    \mathcal{L}_{latent} = \frac{1}{M} \sum_{i=1}^{M} \left \| h_i - \hat{h}_i \right \|^2_2,
\end{equation}
where $\hat{h}_i$ is the $i^{th}$ hidden representation from the Transformer decoder. The overall loss is the sum of the two aforementioned losses:
\begin{equation} 
\label{eq:combined_loss}
    \mathcal{L}=\mathcal{L}_{ce}+\beta \mathcal{L}_{latent},
\end{equation}
where $\beta$ serves as a hyperparameter scaling the latent loss.

\section{EXPERIMENTAL SETTINGS}

\begin{table}[t!]
\caption{Statistics of Sign Language Datasets. NoF refers to the number of frames.} \label{tab:dataset_stats}
\vspace{-0.5em}
\begin{center}
\centering
\scriptsize
\resizebox{\columnwidth}{!}{%
\begin{tabular}{r|c|c|c|c|c}
\hline
\multicolumn{1}{l|}{} &
  \multicolumn{1}{l|}{Train / Valid / Test} &
  \multicolumn{1}{l|}{Max NoF} &
  \multicolumn{1}{l|}{Min NoF} &
  \multicolumn{1}{l|}{Avg NoF} &
  \multicolumn{1}{l}{FPS} \\ \hline
PHOENIX14T \cite{camgoz2018neural}  & 7,096 / 519 / 642 & 475 & 16 & 116 &25  \\
How2Sign \cite{duarte2021how2sign} & 31,128 / 1,741 / 2,322& 2,579 & 32  & 173 &30\\ \hline
\end{tabular}%
}
\end{center}
\vspace{-2em}
\end{table}

\subsection{Datasets}

We evaluated our method using two different sign language datasets: RWTH-PHOENIX-WEATHER-2014T (PHOENIX14T) \cite{camgoz2018neural} and How2Sign \cite{duarte2021how2sign}. Details for each dataset are presented in Tab. \ref{tab:dataset_stats}. PHOENIX14T is a German Sign Language (DGS) dataset from weather forecasts. This dataset contains 8,257 pairs of German and corresponding DGS videos with word-level annotations. How2Sign \cite{duarte2021how2sign} is a large-scale American Sign Language (ASL) dataset that contains 2,500 instructional videos. For PHOENIX14T, where keypoints are not provided, we used OpenPose \cite{cao2019openpose} and skeleton correction model \cite{zelinka2020neural}, following \cite{hwang2021non,saunders2020progressive}.

\subsection{Preprocessing}

During preprocessing, to ensure consistency across all poses, the keypoints were centered and normalized relative to the shoulder joint \cite{yoon2019robots}. This step guarantees that the length from one shoulder to the other is consistently scaled to a value of 1. To further refine the quality of the data, we also implemented a noise frame removal process. This process starts by calculating the differences between consecutive frames in the keypoints, represented as \(X_{\text{diff}} \in \mathbb{R}^{(T-1) \times V \times C}\). Here, \(T\) denotes the number of frames, \(V\) the number of vertices, and \(C\) the feature dimension. Subsequently, we calculated the Euclidean distance for each joint between consecutive frames, resulting in a distance matrix \( D \in \mathbb{R}^{(T-1) \times V} \). We then computed the average distance per frame \( \overline{D} \), and compared this against a predefined threshold \( \theta \). Frames where \( \overline{D} \) exceeds \( \theta \) are identified as noisy and excluded from further processing. Following the removal of noisy frames, we normalized the remaining keypoints to ensure that they fit within the range of \([-1,1]\). This final normalization step is essential for maintaining uniform scaling and positioning of the keypoints. Texts were converted to lowercase and tokenized via Byte-Pare Encoding (BPE). The vocabularly sizes for this encoding were set at 3,000 for PHOENIX14T dataset and 10,000 for How2Sign dataset.

\begin{table}[t!]
\centering
\scriptsize
\caption{Comparison with the previous methods. The best results are in bold, followed by the second-best in underline.}
\vspace{-0.5em}
\begin{center}
\resizebox{\columnwidth}{!}{%
\begin{tabular}{l|ccc|ccc}
\hline
                    & \multicolumn{3}{c|}{PHOENIX14T}                          & \multicolumn{3}{c}{How2Sign} \\             
&FGD$\downarrow$ &DTW-MJE$\downarrow$  & BLEU-4 & FGD$\downarrow$ &DTW-MJE$\downarrow$ & BLEU-4 \\ \hline
PT~\cite{saunders2020progressive}               &360.62	&0.793	&0.59	&391.06	&0.355  &0.57 \\
w/o GN\&FP               &384.23	&0.991	&0.73	&383.20	&0.402   &0.33 \\ \hline
NSLP-G~\cite{hwang2021non} &150.28 &\textbf{0.638}	&5.56	&291.62	&0.327	&0.41 \\
w/o Finetuning &179.40	&0.646	&4.41	&440.95	&0.321	&0.54 \\ \hline
SignVQNet  &\textbf{92.64}	&\underline{0.671}	&\textbf{6.85}	&\underline{82.76}	&\underline{0.319}	&\textbf{0.71} \\ 
w/o Beam search  &\underline{96.60}	&0.670	&\underline{6.39}	&\textbf{81.99}	&\textbf{0.317}	&\underline{0.63} \\ \hline
Ground-truth                  & 0.0  & 0.0  & 8.10  & 0.0  & 0.0  &0.70 \\ \hline
\end{tabular}%
}
\end{center}
\label{tab:eval_sota}
\end{table}

\begin{figure}[t!]
    \centering
    \includegraphics[width=0.95\columnwidth]{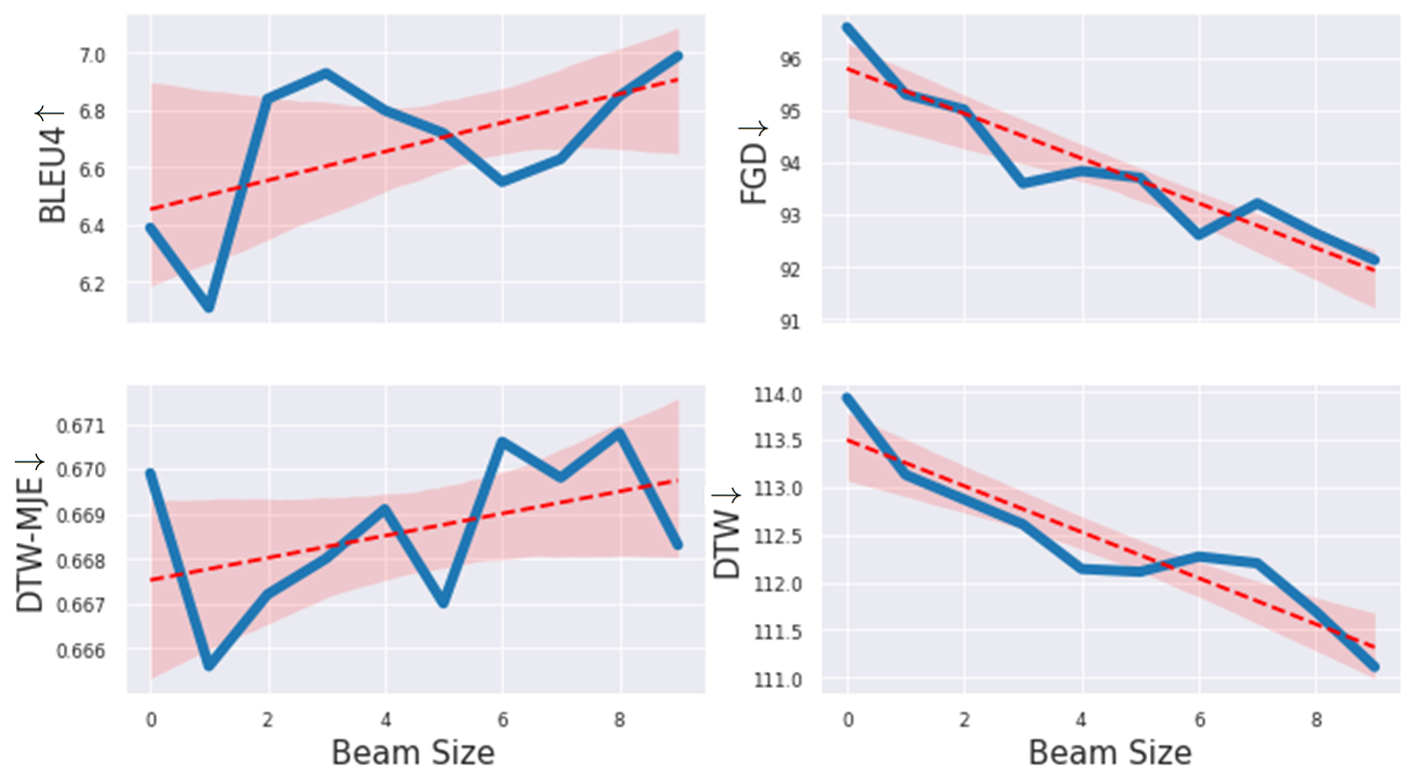}
    \vspace{-0.5em}
    \caption{Performance discrepancy among the SLP metrics in relation to change in beam size. Except for DTW-MJE, all metrics show consistent improvement as beam size increases.}
    \vspace{-1.5em}
    \label{fig:metric_eval}
\end{figure}

\begin{figure*}[t!]
    \centering
    \begin{subfigure}[b]{0.47\linewidth}
    \includegraphics[width=\textwidth]{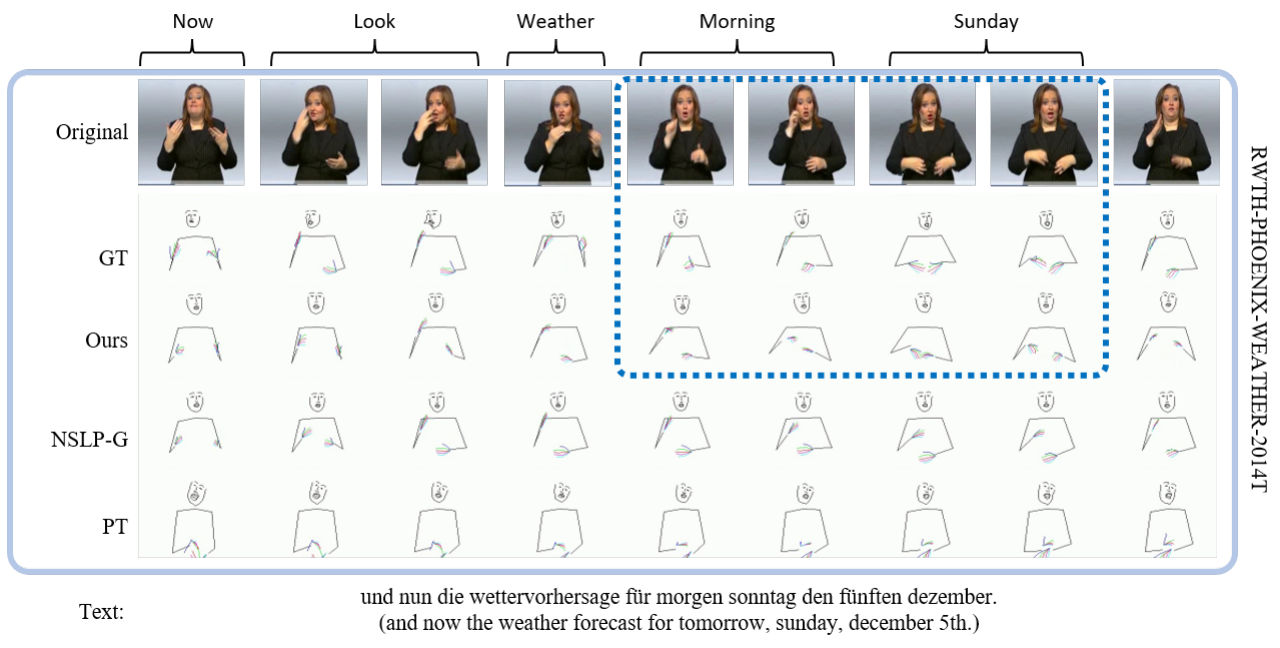}
    \caption{PHOENIX14T}
    \end{subfigure}
    \hfill
    \begin{subfigure}[b]{0.52\linewidth}
    \includegraphics[width=\textwidth]{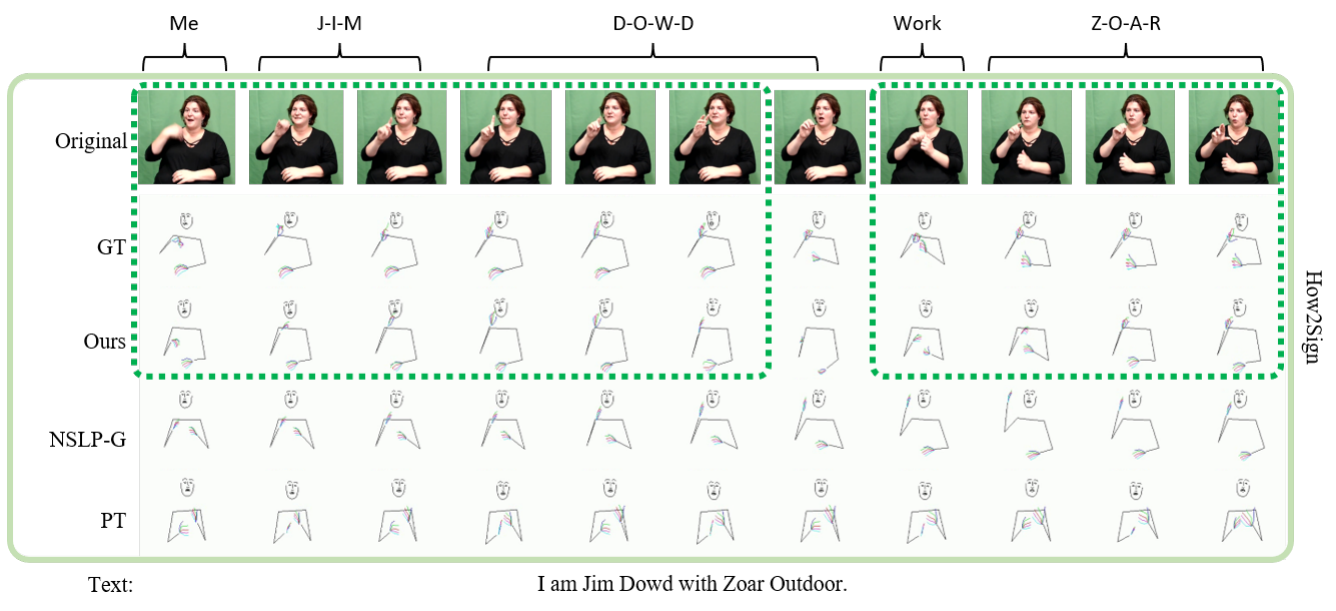}
    \caption{How2Sign}
    \end{subfigure}
    \caption{We present a visual comparison between our method and the baselines on both (a) PHOENIX-2014-T and (b) How2Sign. As highlighted in the dashed boxes, our method generates more realistic and accurate sign pose sequences. Videos are available at \href{http://nlpcl.kaist.ac.kr/~projects/signvqnet}{\textcolor{magenta}{http://nlpcl.kaist.ac.kr/~projects/signvqnet}}}
    \label{fig:qual_eval}
\end{figure*}

\begin{table*}[t!]
\scriptsize
\centering
\caption{Ablation experiments on PHOENIX14T.}
\begin{subtable}[b]{0.49\linewidth}
\resizebox{\columnwidth}{!}{%
\begin{tabular}{c|ccc|ccc}
\hline
\multicolumn{1}{l|}{} & \multicolumn{3}{c|}{DEV} & \multicolumn{3}{c}{TEST} \\
Window Size            & FGD$\downarrow$ &DTW-MJE$\downarrow$ & BLEU-4  &FGD$\downarrow$ &DTW-MJE$\downarrow$   & BLEU-4 \\ \hline
16                   & \underline{42.95}	&0.310	&\underline{6.75}	&\underline{41.45}	&\underline{0.301}	&\underline{6.55} \\
32                  & \textbf{42.04}	&\textbf{0.302}	&\textbf{6.77}	&\textbf{41.43}	&\textbf{0.299}	&\textbf{6.88} \\
64                  & 44.68	&\underline{0.305}	&6.44	&43.21	&0.305	&6.52 \\ \hline
\end{tabular}%
}
\vspace{0.1em}
\caption{Window Size} \label{tab:L-size}
\end{subtable}
\hfill
\begin{subtable}[b]{0.49\linewidth}
\resizebox{\columnwidth}{!}{%
\begin{tabular}{l|ccc|ccc}
\hline
                    & \multicolumn{3}{c|}{DEV}                          & \multicolumn{3}{c}{TEST} \\             
       Loss    &FGD$\downarrow$ &DTW-MJE$\downarrow$   & BLEU-4 & FGD$\downarrow$ &DTW-MJE$\downarrow$  & BLEU-4 \\ \hline
$\mathcal{L}_{ce}$               &\underline{102.87}	&\underline{0.679}	&\underline{6.30}	&\textbf{96.51}	&\underline{0.674}	&\underline{5.94} \\
$\mathcal{L}_{latent}$ & 504.88	&0.755	&0.36	&512.48	&0.741	&0.40 \\
$\mathcal{L}_{ce} + \mathcal{L}_{latent}$      & \textbf{100.85}	&\textbf{0.676}	&\textbf{6.76}	&\underline{96.60}	&\textbf{0.670}	&\textbf{6.39} \\ \hline
\end{tabular}%
}
\vspace{0.1em}
\caption{Loss Type} \label{tab:loss}
\end{subtable}
\vfill
\begin{subtable}[b]{0.50\linewidth}
\resizebox{\linewidth}{!}{%
\begin{tabular}{l|ccc|ccc}
\hline
\multicolumn{1}{l|}{} & \multicolumn{3}{c|}{DEV} & \multicolumn{3}{c}{TEST}\\
Models            &FGD$\downarrow$ &DTW-MJE$\downarrow$   & BLEU-4 &FGD$\downarrow$ &DTW-MJE$\downarrow$  & BLEU-4 \\ \hline
GRU               &433.76	&0.685	&0.49	&444.74	&0.681	&0.61 \\
\quad +Attn               &383.99	&0.697		&0.74	&402.92	&0.700	&0.81 \\ \hline
Transformer &\underline{103.88}	&\underline{0.682}	&\underline{5.33}	&\underline{102.45}	&\underline{0.676}	&\textbf{6.22} \\
\quad +BERT~\cite{devlin-etal-2019-bert} &\textbf{102.87}	&\textbf{0.679}	&\textbf{6.30}	&\textbf{96.51}	&\textbf{0.674}	&\underline{5.94}\\ \hline
\end{tabular}%
}
\vspace{0.1em}
\caption{Architecture Type} \label{tab:architecture}
\end{subtable}
\hfill
\begin{subtable}[b]{0.45\linewidth}
\resizebox{\columnwidth}{!}{%
\begin{tabular}{c|ccc|ccc}
\hline
\multicolumn{1}{l|}{} & \multicolumn{3}{c|}{DEV}                          & \multicolumn{3}{c}{TEST}                          \\
Vocab Size            & FGD$\downarrow$ &DTW-MJE$\downarrow$   & BLEU-4  &FGD$\downarrow$ &DTW-MJE$\downarrow$  & BLEU-4 \\ \hline
512                   & \underline{42.96}	&0.315	&\underline{6.75}	&\underline{41.81}	&0.312	&6.30 \\
1024                  & \textbf{42.04}	&\textbf{0.302}	&\textbf{6.77}	&\textbf{41.43}	&\textbf{0.299}	&\textbf{6.88} \\
2048                  & 44.68	&\underline{0.306}	&6.24	&44.23	&\underline{0.306}	&\underline{6.77} \\
4096                  & 56.74	&0.325		&6.64	&55.13	&0.323		&6.57 \\
8192                  & 55.96	&0.317	&6.30	&55.58	&0.315	&5.89 \\ \hline
\end{tabular}%
}
\vspace{0.1em}
\caption{Codebook Size} \label{tab:codebook_size}
\end{subtable}
\vspace{-2em}
\end{table*}

\subsection{Implementation Details}

For our experiments, we utilized the Gumbel-Softmax relaxation and annealed its temperature, $\tau$, from 0.9 down to 0.1. The parameters $\alpha$ and $\beta$ were set at 0.1 and 0.001, respectively. We used 4 STGP blocks. In configuring the Transformer model, we set the hidden size to 768, with 4 layers and 8 attention heads, a dropout rate to 0.1, and an intermediate size to 1,024. We used the AdamW optimizer \cite{Loshchilov2017DecoupledWD} with a learning rate set at 0.0001. To encode the spoken language setnecne, we used the pre-trained BERT\footnote{\href{https://huggingface.co/models}{\textcolor{magenta}{https://huggingface.co/models}}} (bert-base-cased and bert-base-german-cased), fine-tuned during training. We selected a checkpoint that minimizes the score against FGD metric. The entire training process ran for 500 epochs, taking approximately 24 hours on a Tesla A100 GPU, with a batch size of 64.

\subsection{Evaluation Metrics}

We used a range of evaluation metrics to assess our method. The Back-Translation (BT) \cite{saunders2020progressive} was used to compute BLEU-4 by translating the produced sign pose sequence back into spoken language for comparison with the original text. As a back-translation model, we trained Joint-SLT \cite{camgoz2020sign} on PHOENIX14T and How2Sign, following \cite{huang2021towards,hwang2021non,saunders2020progressive,tang2022gloss}. In addition, we used the DTW-MJE \cite{huang2021towards}, which combines Dynamic Time Warping (DTW) with Mean Joint Error (MJE) to measure discrepancies between predicted and actual sign pose sequences. Additionally, we used Fr\'{e}chet Gesture Distance (FGD) \cite{yoon2020speech} to evaluate the visual fidelity of the generated sign pose sequence by comparing the distributions of real and generated sequences.

\section{EXPERIMENTAL RESULTS}

\subsection{Quantitative Results} \label{sec:sota}

We compared our method with the previous gloss-free SLP methods: PT \cite{saunders2020progressive} and NSLP-G \cite{hwang2021non}. For PT, we employ its base and Gaussian and Future Prediction (GN\&FP) settings. PT was modified to exclude the use of additional ground-truth data, as recommended by \cite{huang2021towards,tang2022gloss}, to ensure a fair comparison. For NSLP-G, we used the frozen and fine-tuning option during training. As shown in Tab. \ref{tab:eval_sota}, our method outperforms the baselines on both datasets, especially in terms of FGD and BLEU-4. On How2Sign, which features longer frame sequences than PHOENIX14T (Tab. \ref{tab:dataset_stats}), our method shows its robustness in generating extended pose sequences. An interesting observation arises from PHOENIX14T. While our method outperforms in FGD and BLEU-4, it lags behind NSLP-G in DTW-MJE. This is mainly due to the unique evaluating manner in DTW-MJE. We delve deeper into this observation in the subsequent sections.

\subsection{Analyzing Evaluation Metrics for SLP}

In the domain of SLP, selecting reliable metrics is crucial for the comprehensive evaluation of the performance of generative models. While the field currently relies on metrics such as FGD, DTW-MJE, and BT, identifying the optimal metric for comprehensive evaluation remains an open question. This challenge is exemplified by the conflicting results observed between our method and NSLP-G, as highlighted in Sec. \ref{sec:sota} and by \cite{Arkushin_2023_CVPR}. Specifically, these discrepancies arise from the differences in loss functions employed by these models. For instance, our method uses CE loss, which focuses on sequential prediction accuracy and preserving the structure of sign language. This emphasis on the sequential structure may affect the model's performance in DTW-MJE, a metric that primarily evaluates the spatial accuracy of keypoints. By contrast, NSLP-G utilizes MSE loss, focusing on the spatial accuracy on a frame-by-frame basis, which typically results in better scores in DTW-MJE.

To further investigate these disparities, we conducted an additional analysis to see how varying beam sizes affects the performance of our model on PHOENIX14T. Our findings, as shown in Fig. \ref{fig:metric_eval}, indicate that FGD and BLEU-4 generally improve with larger beam sizes, whereas DTW-MJE tends to decrease. Additionally, we included DTW in our analysis for a more comprehensive evaluation. It is worth noting that DTW-MJE, by enforcing alignment between the generated and ground-truth sign pose sequences, might not always provide a true comparison. Specifically, DTW aims to minimize the distance by aligning sequences, which may not always reflect the actual temporal alignment. When combined with MJE, this can lead to inconsistent error measurements. This shows the need for careful consideration of the development of new metrics tailored to the specific challenges of SLP evaluation.

\subsection{Qualitative Results}

We offer a visual comparison of the sign pose sequences generated by our method in contrast to the baselines on PHOENIX14T and How2Sign. As shown in Fig. \ref{fig:qual_eval}, our method generates more accurate signs, such as ``Morning'', ``Sunday'', ``Me'', and ``Work'', compared to the baselines. 

\subsection{Ablation Study}

We investigated the effects of various components and design choices of our method on PHOENIX14T, the most widely used dataset in sign language research. Tab. \ref{tab:architecture} shows that the Transformer encoder-decoder model outperforms GRU-based networks, which are commonly used for human motion tasks \cite{gopalakrishnan2019neural,sang2020human,yoon2020speech}. Furthermore, employing the pre-trained BERT \cite{devlin-etal-2019-bert} significantly improved its performance. Regarding the loss functions, a combination of $\mathcal{L}_{ce}$ and $\mathcal{L}_{latent}$ yields the best performance, as shown in Tab. \ref{tab:loss}. Optimal performance was achieved with a window size $L$ set to 32 as shown in Tab. \ref{tab:L-size}, and a codebook size of 1,024 showed the best performance as shown in Tab. \ref{tab:codebook_size}.
 
\section{CONCLUSION}

In this paper, we introduced SignVQNet, gloss-free Sign Language Production (SLP) that leverages vector quantization to derive discrete tokens from sign pose sequences. This enables genuine autoregression without the need for ground-truth data during inference, addressing the shortcomings of the previous autoregressive SLP model. In our experiments, we demonstrate its superiority over prior methods, achieving the state-of-the-art performance on both PHOENIX14T and How2Sign. Additionally, we highlight the reliability of BT and FGD as evaluation metrics, while noting inconsistencies in the DTW-MJE metric.
 
\section{ACKNOWLEDGMENTS}

This work was supported by the Technology Innovation Program (20014406, Development of interactive sign language interpretation service based on artificial intelligence for the hearing impaired) funded by the Ministry of Trade, Industry \& Energy (MOTIE, Korea).



{\small
\bibliographystyle{ieee}
\bibliography{sample_FG2024}
}

\end{document}